# Multiview Subspace Clustering of Hyperspectral Images based on Graph Convolutional Networks

Xianju Li[1,2(✉)], Renxiang Guan[1], Zihao Li[1], Hao Liu[1] and Jing Yang[3]

,[1] Faculty of Computer Science, China University of Geosciences, Wuhan 430074, China
[2] Hubei Key Laboratory of Intelligent Geo-Information Processing, China University of Geosciences, Wuhan 430074, China
[3] Geophysical and Geochemical Exploration Institute of Ningxia Hui Autonomous Region, Yinchuan 750021, China
`{ddwhlxj,grx1126,lizihao,liu_hao}@cug.edu.cn`

**Abstract.** High-dimensional and complex spectral structures make clustering of hyperspectral images (HSI) a challenging task. Subspace clustering has been shown to be an effective approach for addressing this problem. However, current subspace clustering algorithms are mainly designed for a single view and do not fully exploit spatial or texture feature information in HSI. This study proposed a multiview subspace clustering of HSI based on graph convolutional networks. (1) This paper uses the powerful classification ability of graph convolutional network and the learning ability of topological relationships between nodes to analyze and express the spatial relationship of HSI. (2) Pixel texture and pixel neighbor spatial-spectral information were sent to construct two graph convolutional subspaces. (3) An attention-based fusion module was used to adaptively construct a more discriminative feature map. The model was evaluated on three popular HSI datasets, including Indian Pines, Pavia University, and Houston. It achieved overall accuracies of 92.38%, 93.43%, and 83.82%, respectively and significantly outperformed the state-of-the-art clustering methods. In conclusion, the proposed model can effectively improve the clustering accuracy of HSI.

## 1 Introduction

The development of spectral imaging technology has enabled hyperspectral image (HSI) to emerge as an effective tool for detection technology, and has promoted the development of various fields including environmental monitoring [1], geological exploration [2], national defense security [3], and mineral identification [4]. In contrast to traditional color images, HSI possess higher resolution and richer spectral information, and provide more accurate ground object information. Accordingly, various HSI processing technologies have emerged to meet the demands of the times.

Supervised HSI classification methods have achieved significant progress in recent decades, including machine learning models such as support vector machines [5] and deep convolutional neural networks [6,10]. These methods require artificially labeled data as training samples, but labeling HSI pixels is a time-consuming task that demands professional knowledge [8]. To alleviate this burden, and to overcome the limitation of scarce label information in HSI, unsupervised learning, represented by clustering, has garnered extensive attention [9]. HSI clustering enables automatic data processing and interpretation. However, due to the large spectral variability and complex spatial structure of HSI, clustering remains a challenging task [10].

HSI clustering refers to the process of partitioning pixels into corresponding groups based on their intrinsic similarity to facilitate analysis and interpretation. The objective is to ensure that pixels within the same cluster have high intra-class similarity while those across different clusters have low inter-class similarity [11]. Over the years, numerous clustering algorithms have been proposed and widely used in practice. These algorithms include methods based on cluster centers such as k-means clustering [12], and fuzzy c-means clustering (FCM) [13]. Other clustering techniques rely on feature space density distribution, such as the mean shift algorithm [14], and the clustering algorithm based on integrated density analysis [15]. However, these methods are relatively sensitive to initialization and noise, and rely heavily on similarity measures. To gain a deeper understanding of the underlying structure of HSI data, subspace clustering algorithms have received extensive attention and demonstrated remarkable results [34].

The subspace clustering algorithm is a hybrid of traditional feature selection techniques and clustering algorithms, where feature subsets or feature weights corresponding to each data cluster are obtained during the clustering and division of data samples [16]. Sparse subspace clustering [17] and low-rank subspace clustering [18] are representative examples of this algorithm. The crucial aspect of this algorithm is to identify the sparse representation matrix of the original data, construct a similarity graph based on the corresponding matrix, and then obtain the clustering result using spectral clustering. However, these clustering algorithms only analyze the spectral information of HSI, leading to suboptimal results when used for HSI clustering. To address this issue, Zhang et al. proposed a spectral space sparse subspace clustering method ($S_4C$) [19] that leverages the rich spatial environment information carried by HSI in the form of a data cube to improve clustering performance. Literature [20] presents the $l_2$-norm regularized SSC algorithm that integrates adjacent information into the coefficient matrix via $l_2$-norm regularization constraints.

Subspace clustering methods have proven to be effective in numerous applications, but their performance in complex HSI scenes is often limited due to their lack of robustness. Deep clustering models have been proposed to address this limitation by extracting deep and robust features. Pan et al. [24] proposed a deep subspace clustering model using multi-layer autoencoders to learn self-expression. Moreover, deep clustering models based on graph convolutional neural networks (GCN) have gained popularity due to their ability to capture neighborhood information. Zhang et al. [26] introduced hypergraph convolutional subspace clustering, which fully exploits the high-order relationships and long-range interdependencies of HSI. To extract a deep spectral space representation and robust nonlinear affinity, Cai et al. [27] proposed a graph regularized residual subspace clustering network, which significantly improves the clustering accuracy of HSI.

Despite the success achieved by the algorithms described above in improving clustering performance, they suffer from two significant limitations. Firstly, the direct application of these methods on HSI often produces cluster maps with substantial noise due to the limited discriminative information in the spectral domain, the complexity of ground objects, and the diversity of spectral features in the same class [28]. Secondly, these methods have been tested on a single view, and extensive experiments have demonstrated that incorporating complementary information from multiple views can significantly enhance clustering accuracy [29]. To address these limitations, various multi-view clustering techniques have been proposed. For instance, Tian et al. [30] performed multi-view clustering by using multiple views simultaneously but the method is sensitive to noise. Chen et al. [31] applied multi-view subspace clustering to polarized HSI. Huang et al. [32] combined local and non-local spatial information of views to learn a common intrinsic cluster structure, which led to improved clustering performance. Lu et al. [33] proposed a method that combined spectral and spatial information to build capability regions and employed multi-view kernels for collaborative subspace clustering. However, the information weights of different views have not been considered in these methods when fusing information from multiple views, which can lead to the loss of important information.

To address the challenges, we proposed a novel approach to subspace clustering for HSI. The method combines both texture and space-spectral information and leverages graph convolutional networks. By integrating neighborhood information through GCNs and using the information of two views, we aim to achieve effective representation learning. The contributions of this work are threefold:

1) We provide a novel deep multi-view clustering algorithm for HSI clustering, namely MSCGC, which can simultaneously learn texture information and depth spectral spatial information;

2) We use the powerful feature extraction ability of GCN and the learning ability of topological relationships between nodes to analyze and express the spatial relationship of HSI.

3) An attention-based fusion module is adopted to adaptively utilize the affinity graphs of the two views to build a more discriminative graph.

## 2  Related Work

### 2.1  Multi-view subspace clustering

Multi-view clustering techniques have demonstrated significant accomplishments in various domains [35, 36]. In HSI analysis, these methods can be leveraged to enhance the clustering performance by integrating multiple sources of information. Multi-view data $\{X^{(p)}\}_{p=1}^{v}$, where $\mathbf{X}^{(p)} \in \mathbf{R}^{d^{(p)} \times n}$ signifies the data from the $p$th view, with its top dimension $d^{(p)}$. Subspace clustering is premised on the assumption that each data point can be expressed as a linear combination of other points within the same subspace. Based on the above assumptions, the data $X^{(p)}$ of each view itself is used as a dictionary to construct the subspace representation model as:

$$X^{(p)} = X^{(p)}C^{(p)} + E^{(p)}, \quad (1)$$

where $C^{(p)} \in \mathbb{R}^{n \times n}$ is the self-expressive matrix on each view, and $E^{(p)}$ is the representation error. Multi-view subspace clustering methods are usually expressed as follows:

$$\min_{S^{(p)}} \|X^{(p)} - X^{(p)}C^{(p)}\|_F^2 + \lambda f(C^{(p)}),$$
$$\text{s.t. } S^{(p)} \geq 0, C^{(p)^T}\mathbf{1} = \mathbf{1} \quad (2)$$

where $C^{(p)^T}\mathbf{1} = \mathbf{1}$ ensures that each column of $C^{(p)}$ adds up to 1, indicating that each sample point can be reconstructed by a linear combination of other samples. $C_{i,j}^{(p)}$ represents the weight of the edge between the i-th sample and the j-th sample, so $C^{(p)}$ can be regarded as an n×n undirected graph. $f(\cdot)$ is the regularization function, and $\lambda$ is a parameter to balance regularization and loss. Different $f(\cdot)$ bring self-expression matrix $C^{(p)}$ satisfying different constraints. Examples include imposing sparsity constraints on matrices or finding low-rank representations of data.

After obtaining the self-expression matrix $\{C^{(p)}\}_{p=1}^v$ on each view, they are fused to obtain a unified self-expression matrix $C \in \mathbb{R}^{n \times n}$, in some methods this step is also performed together with the learning phase on each view. Taking the consistent C as the input of spectral clustering, the final clustering result is obtained.

### 2.2 Hyperspectral Image Clustering

The analysis of HSI data presents a challenging task, primarily due to the high dimensionality, high correlation, and complex distribution of spectral data. Traditional supervised classification methods demand a considerable number of labeled samples, which may be arduous to obtain. As a solution, unsupervised HSI clustering has emerged as a significant research topic in recent years. The primary objective of HSI clustering is to cluster pixels into distinct classes or clusters based on their spectral characteristics, without any prior knowledge of the data. This technique has proven to be effective in overcoming the limitations of traditional supervised classification methods, as it eliminates the need for labeled samples and can reveal previously unknown relationships in the data.

Among existing HSI clustering methods, subspace clustering has shown promising results due to its robustness. In recent years, graph neural network has attracted people's attention due to its robust feature extraction ability, and its combination with subspace clustering is also a research hotspot. Cai et al. [25] effectively combined structure and feature information from the perspective of graph representation learning and proposed a graph convolutional subspace clustering framework. Zhang et al. [26] proposed hypergraph convolutional subspace clustering to fully exploit the high-order relationships and long-range interdependencies of HSI. To extract deep spectral space representation and robust nonlinear affinity, Cai et al. [27] proposed a graph regularized residual subspace clustering network, which greatly improves the HSI clustering accuracy.

## 3 Method

As shown in Figure 1, the framework proposed in this paper consists of three important parts, namely the multi-view graph construction module, the dual-branch representation modle and the attention fusion module. We first introduce the multi-view graph building blocks, and then introduce the rest sequentially.

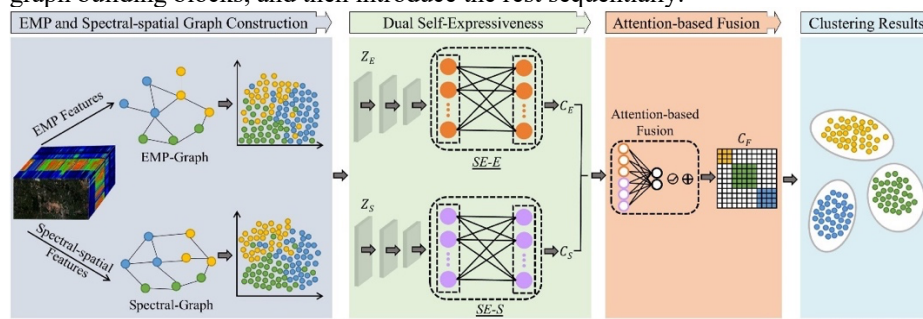

**Fig. 1.** Illustration of the proposed framework, consisting of three important parts, namely the multi-view graph construction module, the dual-branch representation module and the attention fusion module.

### 3.1 EMP and Spectral-spatial Graph Construction

The primary objective of this module is to construct multi-view graphs by extracting texture and spatial spectral features as distinct views. Prior to addressing the primary concern, we must contend with the fact that HSI typically contain a large number of redundant bands. To address this issue, Principal Component Analysis (PCA) is used to reduce the number of bands to d dimensions. Next, the sliding window technique is applied to capture pixel points and their surrounding adjacent pixels, while the patch method is used to represent the data point [25]. Additionally, the EMP algorithm [37] is utilized to corrode the image and retain the texture information. Finally, we obtain two datasets $\{X_1^p, X_2^p\}_{p=1}^n$.

GCN is a neural network that can only operate on data structured as graphs. Therefore, a crucial step in using GCNs is to transform the input data into a graph representation. In this study, we use the K-nearest neighbor (KNN) method to construct a topological graph from the processed data. Specifically, we treat each sample in the data as a node in the graph, and the KNN method is used to determine the connections between nodes. KNN calculates the Euclidean distance between samples within each view and constructs adjacency matrices AE and AS. Finally, by combining the adjacency matrix with the corresponding data, we obtain the EMP map and space spectrum map. This approach facilitates the use of GCNs in multi-view learning tasks by transforming the input data into a graph representation that is amenable to GCN operations.

### 3.2 Graph Convolution Self-Expressive Module

Different from models based on convolutional neural networks, GCN can make full use of the dependencies between nodes and the feature information of each neighbor node through graph convolution operations. The formula for spectrogram convolution is as follows:

$$H^p(l+1) = \sigma\left(\widetilde{D}^{-\frac{1}{2}^p} \widetilde{A}^p \widetilde{D}^{-\frac{1}{2}^p} H^p(l) W^p(l)\right), \qquad (2)$$

where $\widetilde{A}^p = A^p + I_N$, $A^p$ is the adjacency matrix in $p$ view, $I_N$ is the identity matrix; $\widetilde{D}^p$ and $W^p(l)$ are the degree matrix and The weight matrix of layer $l+1$, $\sigma$ is the activation function; $H(l)$ is the data representation of layer $l$, when $l$ is equal to 0, $H(0)$ is the input multi-view data.

Following the graph convolutional subspace clustering defined in [25, 26], we define graph self-expression as:

$$\arg\min_{\mathbf{Z}} \| \mathbf{X}\overline{\mathbf{A}}\mathbf{Z} - \mathbf{X} \|_q + \frac{\lambda}{2} \| \mathbf{Z} \|_q \text{, s.t. diag}(\mathbf{Z}) = 0, \qquad (3)$$

where q denotes the appropriate matrix norm and λ is the trade-off coefficient. $\mathbf{X}\overline{\mathbf{A}}\mathbf{Z}$ can be viewed as a special linear graph convolution operation parameterized by Z, where Z is the self-expressive coefficient matrix and $\overline{\mathbf{A}}$ represents the normalized neighbor matrix.

### 3.3 Attention Fusion Module

After obtaining affinity matrices $Y^p$, we need to fuse them together to construct the final affinity graph, and we will apply spectral clustering to the final affinity matrix.. We utilize an attention-based fusion module to learn the importance $a^p$ of each view as follows:

$$a^p = att(Y^p) \qquad (5)$$

where $a^p \in R^{n \times 1}$ are used to measure the importance of each view. The details of the attention module are provided below. In the first step, we concatenate affinity matrices $Y^p$ as $[Y_1 \cdots Y_p] \in R^{n \times pn}$, and introduce a weight matrix $W \in R^{pn \times p}$ to capture the relationships between the self-expression matrices. Next, we apply the tanh function to the product of $[Y_1 \cdots Y_p]$ and $W$ for a nonlinear transformation. Finally, we use softmax and the $\ell_2$ function to normalize the attention values, resulting in the final weight matrix:

$$a^p = \ell_2(softmax(tanh([Y_1 \cdots Y_p] \cdot W))) \qquad (6)$$

Obtaining the weight matrix can realize the fusion operation, and the fused self-expression matrix $Y_F$ is:

$$Y_F = f(Y_1 \cdots Y_p) = \sum_{i=1}^{N} (a_i 1) \odot Y_i \tag{7}$$

where $1 \in R^{1 \times n}$ is a matrix with all elements being 1, and $\odot$ represents the Hadamard product of the matrix.

## 4 Result

In this section, we show the clustering results of the proposed models on three generic datasets and compare them with several state-of-the-art clustering models.

### 4.1 Set Up

**Datasets:** We have utilized three authentic HSI datasets, namely, Indian Pines, Pavia University, and Houston2013, which were captured using AVIRIS, ROSIS, and ITRES CASI-1500 sensors, respectively. To ensure computational efficacy, we selected a sub-scene from each dataset for the experimental analysis, and the specifications of these sub-scenes are presented in Table 1. Notably, Houston2013, which is a dataset from the 2013 GRSS competition, exhibits considerable diversity. Given that HSI generally comprise numerous superfluous bands, we employed the Principal Component Analysis (PCA) algorithm to reduce the spectral dimensions to four, which encompass at least 96% of the data's variance. Furthermore, to maintain model accuracy while enhancing computational efficiency, we optimized the hyperparameters of the model, and these adjustments are listed in Table 2.

**Baseline**: We use six experimental benchmark methods to compare our model. These include two classic clustering algorithms, SC and SSC, as well as excellent clustering algorithms in recent years, including $l_2$-SSC combined with $l_2$-norm, robust manifold matrix factorization (RMMF), graph convolutional subspace clustering network (EGCSC) and deep spatial spectral subspace clustering (SCNet). Some experimental results refer to other literature [25, 26], and remaining experiments are implemented using Python 3.9 and run on an Intel i9-12900H CPU.

Table 1. Introduction to the datasets information used in the experiment

| Datesets | Indian Pines | Pavia University | Houston-2013 |
|---|---|---|---|
| Pixels | 85×70 | 200×100 | 349×680 |
| Coordinates | 30-115, 24-94 | 150-350, 100-200 | 0-349, 0-680 |
| Channels | 200 | 103 | 144 |
| Samples | 4391 | 6445 | 6048 |
| Clusters | 4 | 8 | 12 |

Table 2. Experiment important hyperparameters setting information

| Datesets | Indian Pines | Pavia University | Houston-2013 |
|---|---|---|---|
| Input size | 13 | 11 | 11 |
| k | 30 | 30 | 25 |
| λ | 100 | 1000 | 1000 |

## 4.2 Quantitative Results

Table 3 illustrates the clustering performance of the proposed MSCGC model and six other models on three benchmark datasets, namely Indian Pines, Pavia University, and Houston2013. The results indicate that the proposed model outperforms the other models in terms of three evaluation metrics: overall accuracy (OA), normalized mutual information (NMI), and Kappa coefficient. Notably, the proposed model exhibits a statistically significant improvement in NMI over the suboptimal method by more than 15% on the Indian Pines and Houston datasets. Furthermore, we can observe the following trends:

**Table 3.** The results of all methods on the experimental data set, the best results of each row are marked in bold

| Dataset | Metric | SC | SSC | $l_2$-SSC | NMFAML | EGCSC | SCNet | MSCGC |
|---|---|---|---|---|---|---|---|---|
| InP. | OA | 0.6841 | 0.4937 | 0.6645 | 0.8508 | 0.8483 | 0.8914 | **0.9238** |
| | NMI | 0.5339 | 0.2261 | 0.3380 | 0.7264 | 0.6442 | 0.7115 | **0.8925** |
| | Kappa | 0.5055 | 0.2913 | 0.5260 | 0.7809 | 0.6422 | 0.8413 | **0.8791** |
| PaU. | OA | 0.7691 | 0.6146 | 0.5842 | 0.8967 | 0.8442 | 0.9075 | **0.9343** |
| | NMI | 0.6784 | 0.6545 | 0.4942 | 0.9216 | 0.8401 | 0.9386 | **0.9394** |
| | Kappa | 0.8086 | 0.4886 | 0.3687 | 0.8625 | 0.7968 | 0.8777 | **0.9265** |
| Hou. | OA | 0.3661 | 0.5526 | 0.4228 | 0.6346 | 0.6238 | 0.7426 | **0.8382** |
| | NMI | 0.5067 | 0.7531 | 0.5167 | 0.7959 | 0.7754 | 0.7567 | **0.8981** |
| | Kappa | 0.2870 | 0.5022 | 0.3609 | 0.5910 | 0.5812 | 0.7138 | **0.8329** |

1) Introducing deep learning methods and regularization into clustering can better improve accuracy. Methods based on deep learning, such as EGCSC and SCNet, have greatly improved compared with traditional models. In addition, $l_2$-SSC introduced L2-norm regularization in the traditional model improves the accuracy by 10% on the Indian dataset compared to the SSC method. The introduction of graph regularization by EGCSC has also achieved great improvement.

2) The introduction of multi-view complementary information is beneficial to improve the clustering accuracy. NMFAML combined with homogeneous information is also used to extract feature information in comparative experiments. Similarly, our model combines texture information and uses GCN to aggregate neighborhood information, which effectively improves the clustering. It achieved 92.38%, 93.43% and 83.82% accuracy respectively on the three datasets.

3) The experimental accuracy of MSCGC on the three data sets is better than that of EGCSC. Specifically, compared with EGCSC, our model improves by 7.55%, 9.01% and 21.44% respectively on all data sets. This shows that the introduction of multi-view information and the attention fusion module can improve the clustering accuracy very well. So our model sheds new light on HSI clustering.

## 4.3 Qualitative Comparison of Different Methods

Figure 2-4 shows the visualization results of various clustering methods on the Indian, Pavia and Houston2013 datasets. Part (a) of each picture is the true value of the corresponding data set to remove the irrelevant background, and the color of the same class may be different in different methods. Although the SSC algorithm can discover the low-dimensional information of the data from the high-dimensional structure of the data, it does not consider the space constraints. Moreover, traditional subspace clustering methods are difficult to prepare for modeling HSI structures and are weaker than deep learning-based methods to a certain extent. Our model achieves state-of-the-art visualization results on all datasets. Due to the efficient aggregation of texture information, MSCGC has the least salt and pepper noise on the clustering map compared to other methods. At the same time, because the GCN model can aggregate the information of neighbor nodes, it maintains better homogeneity in similar object areas and the boundary of the image remains relatively complete.

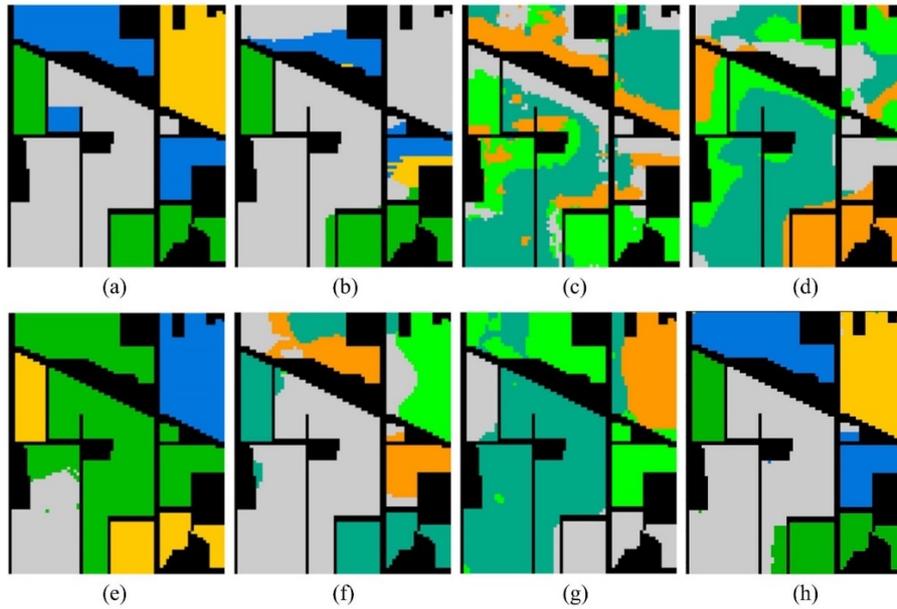

**Fig. 2.** Clustering results obtained by different methods on the Indian Pines dataset. (a) Ground truth, (b) SC 68.41%, (c) SSC 49.37%, (d) L2-SSC 66.45%, (e) NMFAML 85.08%, (f) EGCSC 84.67%, (g) SCNet 89.14%, and (h) MSCGC 92.38%

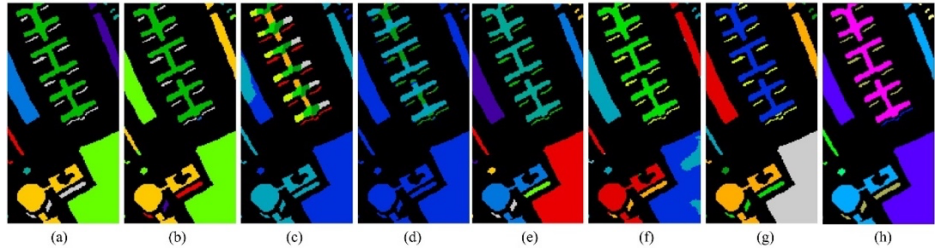

**Fig. 3.** Clustering results obtained by different methods on the Pavia University dataset. (a) Ground truth, (b) SC 76.91%, (c) SSC 64.46%, (d) L2-SSC 58.42%, (e) NMFAML 89.67%, (f) EGCSC 83.79%, (g) SCNet 90.75%, and (h) MSCGC 93.43%.

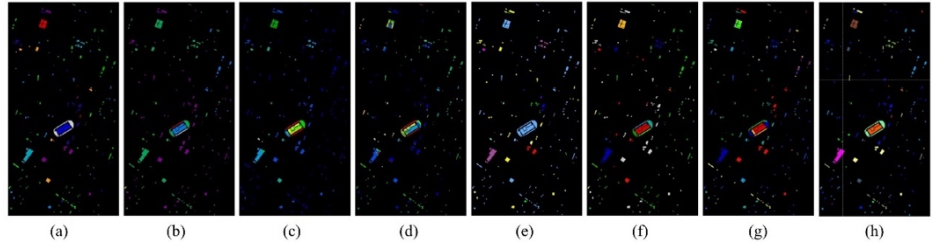

**Fig. 4.** Clustering results obtained by different methods on the Houston2013 dataset: (a) Ground truth, (b) SC 36:61%, (c) SSC 55:26%, (d) '2-SSC 42:28%, (e) NMFAML 63:46%, (h) EGCSC 62:38%, (i) SCNet 74.26%, and (j) GR-RSCNet 83:82%.

### 4.4 Visualization of the Learned Affinity Matrix

We demonstrate the learned affinity matrices on three datasets and visualize them in Fig. 5. From the three figures, it can be seen that the affinity matrix learned by MSCGC is not only sparse, but also has an obvious block diagonal structure, which shows that our model can better identify the internal relationship of clusters, so as to realize accurate HSI classification.

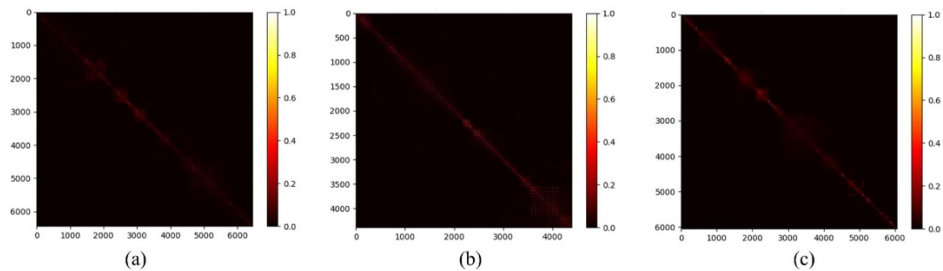

**Fig. 5.** Visualization of deep affinity matrix learned on (a) Indian Pines, (b) Pavia University, and (c) Houston2013 datasets.

## 5      Conclusion

We construct a novel multi-view subspace clustering network for HSI clustering, named MSCGC. Specifically, we first extract the texture information and spatial spectral information of the HSI to construct a multi-view map result. Then, a GCN is used to aggregate neighborhood information and the extracted features are fed into a self-expressive network to learn an affinity graph. We then employ an attention-based strategy to fuse the affinity graphs obtained from the two learning branches. We test our proposed model on three general-purpose HSI datasets and compare it with various state-of-the-art models. The experimental results show that MSCGC has achieved the optimal clustering results, and achieved clustering accuracies of 92.38%, 93.43% and 83.82% on the Indian Pines, Pavia University and Houston2013 datasets, respectively.

The self-expression layer determines that our model is difficult to train on large-scale data sets, and minBatch also has an inhibitory effect on the accuracy of the graph neural network. In the future, we will try to explore the potential of MSCGC on large-scale datasets.

**Acknowledgments:** This work was supported by Natural Science Foundation of China (No. U21A2013 and 42071430), Natural Science Foundation of Ningxia Hui Autonomous Region (2021AAC03453), Opening Fund of Key Laboratory of Geological Survey and Evaluation of Ministry of Education (Grant Number: GLAB2020ZR14 and CUG2022ZR02) and the Fundamental Research Founds for National University, China University of Geosciences (Wuhan) (No. CUGDCJJ202227).